%% file: conference_101719.tex
\documentclass[conference]{IEEEtran}
\IEEEoverridecommandlockouts
\usepackage{cite}
\usepackage{amsmath,amssymb,amsfonts}
\usepackage{algorithmic}
\usepackage{graphicx}
\usepackage{textcomp}
\usepackage{soul}
\usepackage{marvosym}
\usepackage { svg }
\usepackage{mathtools}
\usepackage{multicol}
\usepackage{colortbl}
\usepackage{cite}
\usepackage[]{xcolor}
\usepackage{float}
\usepackage{algorithmic}
\usepackage{subfigure}
\usepackage{graphicx}
\usepackage{textcomp}
\usepackage{caption}
\usepackage[numbers,sort&compress]{natbib}
\usepackage{balance}
\usepackage{listings}
\usepackage{color}
\usepackage{amsmath,amssymb,amsfonts}
\usepackage{algorithmic}
\usepackage{soul}
\usepackage{svg}
\usepackage{pifont}
\usepackage{booktabs}
\usepackage{makecell}
\usepackage{multirow}
\usepackage{algorithm}
\usepackage { xcolor }
\def \BibTeX { { \rm B \kern -.05em { \sc i \kern -.025em b } \kern -.08em
    T \kern -.1667em \lower .7ex \hbox { E } \kern -.125 emX } }
\begin{document}

\title{AMD: Adaptive Masked Distillation for Object Detection}

\author{
    \IEEEauthorblockN{Guang Yang$^{a}$, Yin Tang$^{b}$, Jun Li$^a$\textsuperscript{\Letter},Jianhua Xu$^{a}$,Xili Wan$^{b}$}
    \IEEEauthorblockA{$^a$ School of Computer and Electronic Information, Nanjing Normal University, Nanjing, China}
    \IEEEauthorblockA{$^b$ School of Computer Science and Technology, Nanjing Tech University, Nanjing, China}
    \IEEEauthorblockA{\{Jun Li\}lijuncst@njnu.edu.cn}
 
}

\maketitle

\begin{abstract}
As a general model compression paradigm, feature-based knowledge distillation allows the student model to learn expressive features from the teacher counterpart. In this paper, we mainly focus on designing an effective feature-distillation framework and propose a spatial-channel adaptive masked distillation (AMD) network for object detection. More specifically, in order to accurately reconstruct important feature regions, we first perform attention-guided feature masking on the feature map of the student network, such that we can identify the important features via spatially adaptive feature masking instead of random masking in the previous methods. In addition, we employ a simple and efficient module to allow the student network channel to be adaptive, improving its model capability in object perception and detection. In contrast to the previous methods, more crucial object-aware features can be reconstructed and learned from the proposed network, which is conducive to accurate object detection. The empirical experiments demonstrate the superiority of our method: with the help of our proposed distillation method, the student networks report 41.3\%, 42.4\%, and 42.7\% mAP scores when RetinaNet, Cascade Mask-RCNN and RepPoints are respectively used as the teacher framework for object detection, which outperforms the previous state-of-the-art distillation methods including FGD and MGD.
\end{abstract}


\input{Intro}

\input{Relate}

\input{Approach}

\input{Experiment}

\input{Conclu}

\balance
\bibliographystyle{IEEEtran}
\bibliography{scmd}
\end{document}

%% file: Intro.tex
\section{Introduction}
Recent years have witnessed successful and pervasive applications of Deep Convolutional Neural Networks (CNNs) in various computer vision tasks. However, deep CNNs usually cost a huge amount of computational resources in pursuit of higher performance, which adversely affects their deployment in practical applications and leads to severe parameter redundancy. 
\begin{figure}[htb]
\centerline{\includegraphics[width=0.4\textwidth]{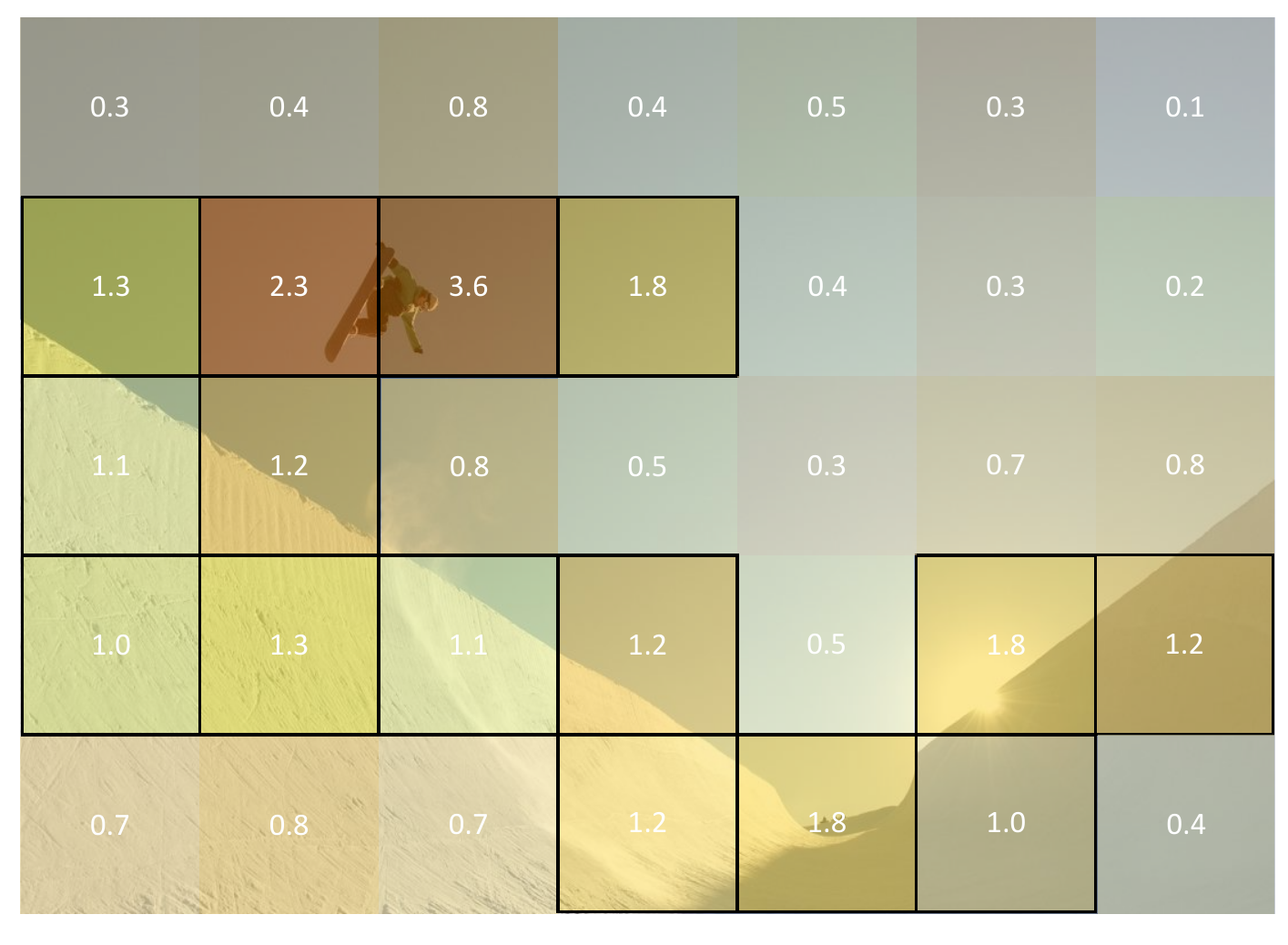}}
\caption{Different regions are quantified with varying attention scores in the feature map of teacher model. The regions with higher scores encode the region importance and should outweigh the low-score regions in the feature masking.}
\label{scores}
\end{figure}
It is therefore necessary to transfer the dark knowledge learned in the complex networks (teacher) to another lightweight network (student). This is also termed as knowledge distillation \cite{hinton2015distilling} which allows the student model to generate expressive features learned from the teacher model. Thus, it is more preferable to deploy the student model with compact network architecture sacrificing minimal loss of performance.

The earliest distillation algorithms function mainly at the output head. The representative examples include logit-based distillation for classification and head-based distillation for detection \cite{chen2017learning}. Recently, a more common distillation strategy emerges as feature-based distillation mechanism. Since only the head or projector after the generated feature varies within different networks, the feature-based distillation approaches can potentially be employed in a variety of tasks. Therefore, it has become a prominent line of research for both model compression and performance improvement due to its simplicity and efficacy.
\begin{figure*}[tb]
\centerline{\includegraphics[width=1.0\textwidth]{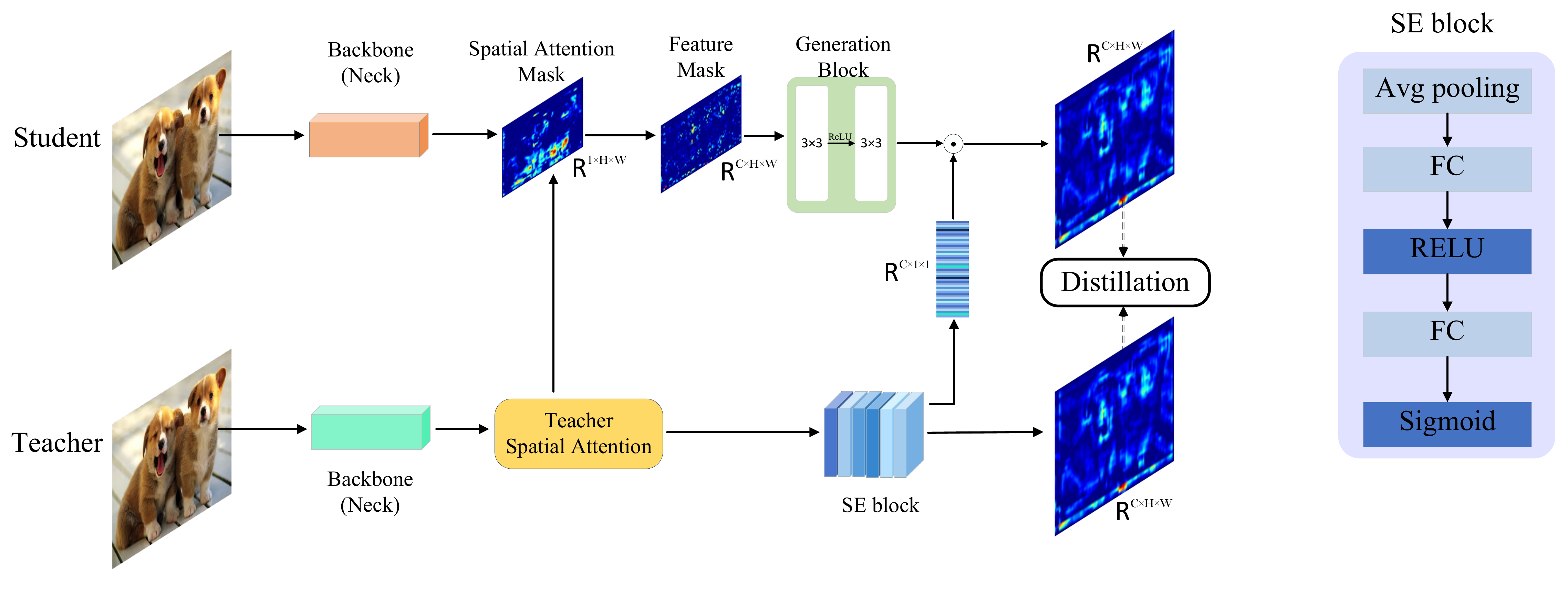}}
\caption{The proposed AMD distillation framework. It first learns the adaptive Region-of-Interest (RoI) via attention-guided feature masking, generating the spatial mask clue from the teacher model imposed on the student feature. Furthermore, we apply the simple and efficient SE layer to the feature of the teacher model, leading to the channel adaptive clues. The auxiliary clues are then fused with the output from the generation block via a Hadamard product, such that the generated feature from the student model is channel adaptive.}
\label{amg}
\end{figure*}
In object detection, in particular, a variety of feature-based distillation approaches have been developed. The earlier research, such as FitNet \cite{romero2014fitnets}, performs distillation at the global level. FGFI \cite{wang2019distilling} operates by distilling the features of high IoU between ground truth and anchors. FGD \cite{yang2022focal} was developed to separate distillation of foreground and background.
Recent research suggests it is preferable for the student model to reconstruct and learn expressive features from the teacher model in the first place instead of following the teacher for generating competitive representations. For instance, MGD \cite{yang2022masked} was proposed to randomly mask pixels in the feature map of student network, leading to reconstructed features of the teacher model via a simple block.

Although MGD further improves the feature distillation by reconstructing the features of masked areas, the masked regions are generated in a random manner. This random operation fails to identify the region-specific importance, and is likely to cause the student model to generate features of the teacher in unimportant regions. As illustrated in Fig. \ref{scores}, the importance of different regions in the feature map of a teacher model can be quantified using the region-specific attention scores. Only the regions with higher scores play critical role in feature masking while the low-score regions should be downplayed. 

To alleviate the above-mentioned drawback, we propose an adaptive masked distillation (AMD) framework which enjoys object-aware spatial and channel adaptivity. On the one hand, we perform attention-guided spatial masking instead of random masking on the feature map of the student network. More specifically, we first learn a spatial attention map from the feature map of the teacher model, producing a region-specific mask. Then, the feature of the student network is adaptively masked by using this attention map. Benefiting from this selective feature masking, it allows subsequent generation block to focus on those adaptively masked important areas, leading to robust and expressive representations. On the other hand, to further explore the object-awareness capability, we leverage a simple and effective SE layer \cite{hu2018squeeze} for modeling the channel attention of the resulting feature of the teacher model. The learned clue and the output from the generation block of students will be fused via a Hadamard product, achieving desirable object-aware channel adaptivity.

To summarize, the contributions of this paper are threefold.
\begin{itemize}
\item First, we develop a spatially adaptive feature masking mechanism for the student model, such that the region-specific importance can be encoded in the features reconstructed and learned from the teacher network.

\item Second, we further explore the channel adaptivity by introducing a simple and efficient SE module to improve the object-aware capability of the student model. 

\item Third, we evaluate our proposed feature distillation network AMD using various detection frameworks including one-stage detector RetinaNet \cite{lin2017focal}, two-stage detector Faster-RCNN \cite{ren2015faster}, and anchor free model RepPoint \cite{yang2019reppoints}. Extensive experimental results demonstrate that our method can help to learn features with sufficient descriptive capability and achieve significant performance gains over the previous state-of-the-art methods.
\end{itemize}

The remainder of this paper is structured as follows. After reviewing the related work in Section II, we elaborate on our method in Section III. Next, we conduct extensive experimental evaluations in Section IV before the paper is finally concluded in Section V.

%% file: Relate.tex
\section{Related Work}
In this section, we comprehensively review the recent advance in object detection and knowledge distillation, both of which are closely related to our method.
\subsection{Object Detection}
As one fundamental vision task, object detection aims to determine the category and location of the objects in an image. Over recent years, the success of CNNs has enormously advanced the research in object detection. In general, the detectors based on deep CNNs can be classified into three categories including anchor-based detectors \cite{ren2015faster,fu2017dssd}, anchor-free detectors \cite{tian2019fcos} and end-to-end detectors \cite{carion2020end}. In particular, anchor-based detection models are divided into two-stage \cite{gidaris2015object,ren2015faster,kong2016hypernet,he2017mask} and one-stage detectors \cite{redmon2018yolov3,ge2021yolox,fu2017dssd}. The former detection method, represented by R-CNN like \cite{girshick2015fast,ren2015faster} algorithms, has a higher detection accuracy, whereas its inference speed is usually unsatisfactory due to expensive computational costs incurred by region proposal network (RPN). As a result, it is impractical for some real-time scenarios. In contrast, one-stage detectors directly perform classification and regression on the anchors without generating proposals beforehand. Thus, they run faster with guaranteed detection performance. 
\begin{figure*}[htb]
\centerline{\includegraphics[width=0.8\textwidth]{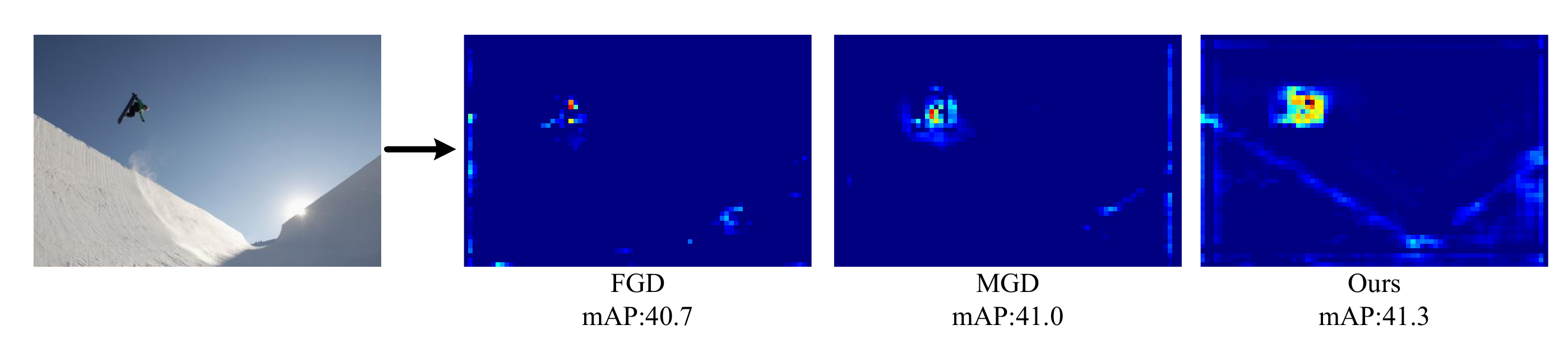}}
\caption{Visualisation of the feature maps obtained by different distillation methods. Teacher detector is RetinaNet-ResNeXt101 while student detector is RetinaNet-ResNet50.}
\label{vis}
\end{figure*}

\par While recent deep networks achieve high detection accuracy, they usually rely on complex backbone structure and significant computational resources \cite{liu2021swin,carion2020end,zhang2022dino,zhu2020deformable}. In this sense, designing lightweight and efficient backbone networks has emerged as a major line of research in object detection. In particular, knowledge distillation, which can transfer sufficient descriptive power from a large network to a small network, is beneficial for designing lightweight backbone with maintained performance close to the large network.
\subsection{Knowledge Distillation}
Recently, knowledge distillation has received increasing attention in model compression, since it is capable of retaining compact model structure with promoted performance. Hinton et al. \cite{hinton2015distilling} first came up with the concept of knowledge distillation by introducing the soft label of the teacher network as part of the loss of the student network, allowing the student network to learn probability distribution fitting of the teacher model for classification task. Moreover, Romero et al. \cite{romero2014fitnets} demonstrated that semantic information in the intermediate layer can also be learned as dark knowledge by student networks. Thus, knowledge distillation can therefore be widely applied to a wide range of downstream tasks. Chen et al. \cite{chen2017learning} distilled the neck feature, classification head, and regression head by setting up three loss functions, respectively. Tang et al. \cite{tang2019learning} carefully designed the distillation weights and distillation loss functions such that they are automatically adjusted between samples for the single-stage object detector. Li et al. \cite{li2017mimicking} used region proposals of the larger network to help the smaller network learn higher semantic information. Zheng et al. \cite{zheng2022localization} transferred the knowledge distillation of the classification head to the location head of object detection, leading to a new distillation mechanism termed Localization Distillation (LD). LD makes logit mimicking become a better alternative to feature imitation, and reveals the knowledge of object category and object location should be handled separately. Dai et al. \cite{dai2021general} developed GID framework which selects distillation areas based on differences between the student and teacher networks. Yang et al. proposed FGD  \cite{yang2022focal} which separates the foreground and background, enabling the student model to learn from the teacher network areas of interest and global knowledge via local and global distillation respectively. Besides, MGD \cite{yang2022masked} imposes random masking on the feature map of the student model, and then generates the feature map reconstructing from the teacher network. However, the uncertainty of random masking may introduce additional noise, producing biased feature map with compromised representation capability.

%% file: Approach.tex
\section{The Proposed Approach}
Recently, a massive amount of distillation methods are carefully designed for various model architectures and tasks. Typically, the feature maps used for distillation usually have high-level semantics and spatial information about adjacent pixels. Therefore, learning these features from the teacher model can significantly improve the performance of the student model. Mathematically, basic feature distillation can be formulated as:
\begin{equation}
L_{f e a}=\frac{1}{C H W} \sum_{k=1}^{C} \sum_{i=1}^{H} \sum_{j=1}^{W}\left(F_{k, i, j}^{T}-f\left(F_{k, i, j}^{S}\right)\right)^{2}
\end{equation}
where $C$, $H$, and $W$ denote the channel, height, and width of the feature map, respectively. $F^{T}$ and $F^{S}$ denote the feature generated from the teacher model and its counterpart from the student model. $f$ represents the adaptation layer that aligns the shape of $F^{S}$ and $F^{T}$. 

Recent research suggests learning and reconstructing the features of the teacher model is a desirable alternative to feature imitation \cite{yang2022masked}. More specifically, expressive features can be generated from the masked regions on the feature map of the student network. However, previous state-of-the-art method mainly performs random feature masking without identifying the importance of different regions on the feature map. In this paper, we attempt to make the student model generate features corresponding to the important areas on the feature map of the teacher network. Towards this end, we propose a spatial-channel adaptive masked distillation strategy termed AMD. In contrast to the random masking strategy in the previous method, we perform feature masking via region-aware attention for identifying the important areas in the feature map of the teacher network. In order to improve the object-aware capability, we further introduce a simple and efficient SE module such that the resulting features are channel adaptive. The framework of our proposed method is illustrated in Fig. \ref{amg}.
\subsection{Spatially adaptive feature masking}
Using random pixels to recover the complete feature map, MGD allows the masked features of the student model to generate features of the teacher model. Thus, it is beneficial for the student network to obtain a better representation. However, the region-specific importance is discarded due to the random masking in MGD. To alleviate this drawback, we carefully design the region-aware feature masking with the help of spatial attention. 
To begin with, we calculate the absolute mean value of the teacher network along the channel dimension:
\begin{equation}
      {G}^{S}(F)=\frac{1}{C}  \sum_{k=1}^{C}\left|F^{T}_{k}\right| 
\end{equation}
where $C$ denotes the channel number of the feature. $F^{T}$ is the feature of the teacher. ${G}^{S}(F)$ is the spatial representation map. Then, the spatial attention mask resulting from the teacher model can be formulated as:
\begin{equation}
 \mathcal {A}^{S}(F)=H \cdot W \cdot \operatorname{softmax}\left({G}^{S}(F) / \mathcal{T}\right)
\end{equation}
where $\mathcal{T}$ is a hyper-parameter introduced in \cite{hinton2015distilling} to change the probability distribution such that the shape of the resulting $\mathcal{A}^{S}$ is $1 \times H \times W$. The attention score for each location represents the level of interest in the teacher network. Furthermore, the mask value is set to 0 when the attention score is greater than $\lambda$ and the rest are set to 1. This can be expressed as:
\begin{equation}
\label{M}
    M_{i, j}=\left\{\begin{array}{ll}
0, & \text { if }\mathcal{A}^{S}_{i, j}>\lambda \\
1, & \text { Otherwise }
\end{array}\right.
\end{equation}
where $\mathcal{A}^{S}_{i, j}$ is the spatial attention score at the point with coordinates $(i,j)$ on the feature map of the teacher network. $\lambda$ is a hyper-parameter to control the number of pixels in the mask. Next, we cover the feature map of the student model with the mask $M$, which can be formulated as follows:
\begin{equation}
    F^{S}_{mask} = F^{S}  \cdot M
\end{equation}

In a nutshell, with the help of this attention-guided feature masking, we can mask out the student feature map according to the important regions of interest on the teacher counterpart, and the resulting feature will contain more important semantic information.

\subsection{Channel adaptive clues generation}
Different from single-object recognization tasks such as image classification, object detection is a dense prediction task focusing on detecting multiple objects. Except for the effective receptive field (ERF), the capability of capturing the object information in different scales can also bring a significant performance fluctuation for a detector, which is not considered in the previous work \cite{carion2020end,yang2022masked,yang2022focal}. Therefore, we utilize a simple and lightweight SE layer \cite{hu2018squeeze} to learn the channel adaptive clue from the teacher feature. The resulting channel adaptive clue will be applied to enhance the student's feature, and further improve the object-awareness capability:
\begin{align}
\notag
    &  F_{clue}^{T}=\sigma\left(\mathbf{W}_{L_{1}}\left(\mathbf{W}_{L_{2}}\left(F^{T}_{avg};\theta_{1}\right);\theta_{2}\right)\right), \\
    &  \mathcal{G}^{S}(F^{S}_{mask})=\mathbf{W}_{C_{1}}\left(ReLU\left(\mathbf{W}_{C_{2}}(F^{S}_{mask};\theta_{1})
    \right)
    ;\theta_{2}\right)\odot  F_{clue}^{T},
\end{align}
where $ F^{T}_{clue}\in \mathbf{R}^{1\times 1\times C}$ denotes the learned channel adaptive clue for the student feature. It is fused with the output from the generation block via a Hadamard product denoted as $\odot$. The $\mathbf{W}_{L}(\cdot;\theta)$ and $\mathbf{W}_{C}(\cdot;\theta)$ are weight matrices of linear projection and convolution layer for SE and generation modules, respectively.

Benefiting from this design, our model further explores the object-aware potential, resulting in a significant improvement over those vanilla counterparts, \textit{i.e.,} models with no channel-adaptive design. More interestingly, we observe that our AMD can achieve a remarkable mAP improvement in the case of detecting \emph{small} objects, demonstrating the effectiveness of our proposed method. We also provide the visualization results of the feature map derived from different distillation models as shown in Fig \ref{vis}. It can be easily observed that the object feature produced from our AMD is more distinguishable than those of methods.
\subsection{Loss function}
Based on the proposed distillation method, we design the following distillation loss for AMD:
\begin{equation}
    \mathcal L_{f e a}= \sum_{k=1}^{C} \sum_{i=1}^{H} \sum_{j=1}^{W} \left(F_{k, i, j}^{T}-  \mathcal{G}^{S}(F^{S}_{mask})\right)^{2}
\end{equation}
where $C$, $H$, and $W$ respectively denote the channel number, height and width of the feature map. $F^{S}_{mask}$ denotes the masked student feature map. Thus, the overall loss function is as follows:
\begin{equation}
    \mathcal{L}_{\text {overall }}\left(F^{{T}}, F^{{S}}\right)=\alpha \cdot \mathcal{L}_{fea}+\mathcal{L}_{original}
\end{equation}
where $\alpha$ is a hyper-parameter to balance distillation loss and original loss, and $\mathcal{L}_{original}$ is the original loss of the detection task.

%% file: Experiment.tex
\section{Experiment}
\subsection{Experimental Setting}
To verify the effectiveness of our AMD for object detection, we evaluate our method on MS COCO2017 \cite{lin2014microsoft} benchmark dataset, which contains 80 object categories and over 160k images. We use 120k training images for training and 5k validation images for testing. For performance measures, we use Average Precision (AP) and Average Recall (AR) to evaluate the performance of different object detectors. Three mainstream detectors including the anchor-based one-stage detector RetinaNet \cite{lin2017focal}, the two-stage detector Faster-RCNN \cite{ren2015faster}, and the anchor-free detector RepPoint \cite{yang2019reppoints} are involved in our comprehensive experiments. In addition, ResNeXt101 and ResNet50 are respectively used as the backbone of the teacher network and its student counterpart. 

\begin{table*}[tb]
\centering
\renewcommand{\arraystretch}{1.1}
\caption{Comparison of our method with other distillation methods for object detection on COCO.}
\resizebox{0.8\textwidth}{!}{%
\begin{tabular}{c|l|llll|llll}
\hline
Teacher                                                                                             & \multicolumn{1}{c|}{Student} & mAP                          & $AP_{S}$                          & $AP_{M}$                          & $AP_{L}$                           & mAR                          & $AR_{S}$                           & $AR_{M}$                          & $AR_{L}$                           \\ \hline
                                                                                                    & RetinaNet-Res50              & 37.4                                        & 20.6                                  & 40.7                                  & 49.7                                  & 53.9                                        & 33.1                                  & 57.7                                  & 70.2                                  \\
                                                                                                    & FKD \cite{zhang2020improve}                          & 39.6 (+2.2)                                  & 22.7                                  & 43.3                                  & 52.5                                  & 56.1 (+2.2)                                  & 36.8                                  & 60.0                                  & 72.1                                  \\
                                                                                                    & FGD \cite{yang2022focal}                           & 40.7 (+3.3)                                  & 22.9                                  & 45.0                                  & 54.7                                  & 56.8 (+2.9)                                  & 36.5                                  & 61.4                                  & 72.8                                  \\
                                                                                                    & MGD \cite{yang2022masked}                          & 41.0 (+3.6)                                  & 23.4                                  & 45.3                                  & \textbf{55.7}                         & 57.0 (+3.1)                                  & 37.2                                  & 61.7                         & 72.8                                  \\
\multirow{-5}{*}{\begin{tabular}[c]{@{}c@{}}RetinaNet\\ ResNeXt101\\ (41.0)\end{tabular}}           & \cellcolor[HTML]{EFEFEF}AMD (ours) & \cellcolor[HTML]{EFEFEF}\textbf{41.3 (+3.9)} & \cellcolor[HTML]{EFEFEF}\textbf{23.9} & \cellcolor[HTML]{EFEFEF}\textbf{45.4} & \cellcolor[HTML]{EFEFEF}\textbf{55.7} & \cellcolor[HTML]{EFEFEF}\textbf{57.4 (+3.5)} & \cellcolor[HTML]{EFEFEF}\textbf{38.2} & \cellcolor[HTML]{EFEFEF}\textbf{61.7} & \cellcolor[HTML]{EFEFEF}\textbf{73.5} \\ \hline
                                                                                                    & RepPoints-Res50              & 38.6                                        & 22.5                                  & 42.2                                  & 50.4                                  & 55.1                                        & 34.9                                  & 59.4                                  & 70.3                                  \\
                                                                                                    & FKD \cite{zhang2020improve}                          & 40.6 (+2.0)                                  & 23.4                                  & 44.6                                  & 53.0                                  & 56.9 (+1.8)                                  & 37.3                                  & 60.9                                  & 71.4                                  \\
                                                                                                    & FGD \cite{yang2022focal}                          & 42.0 (+3.4)                                  & 24.0                                  & 45.7                                  & 55.6                                  & 58.2 (+3.1)                                  & 37.8                                  & 62.2                                  & 73.3                                  \\
                                                                                                    & MGD \cite{yang2022masked}                          & 42.3 (+3.7)                                  & 24.4                                  & 46.2                                  & 55.9                                  & 58.4 (+3.3)                                  & 40.4                                  & 62.3                                  & 73.9                                  \\
\multirow{-5}{*}{\begin{tabular}[c]{@{}c@{}}RepPoints\\ ResNeXt101\\ (44.2)\end{tabular}}           & \cellcolor[HTML]{EFEFEF}AMD (ours) & \cellcolor[HTML]{EFEFEF}\textbf{42.7 (+4.1)} & \cellcolor[HTML]{EFEFEF}\textbf{24.8} & \cellcolor[HTML]{EFEFEF}\textbf{46.5} & \cellcolor[HTML]{EFEFEF}\textbf{56.3} & \cellcolor[HTML]{EFEFEF}\textbf{58.8 (+3.7)} & \cellcolor[HTML]{EFEFEF}\textbf{40.6} & \cellcolor[HTML]{EFEFEF}\textbf{62.4} & \cellcolor[HTML]{EFEFEF}\textbf{74.1} \\ \hline
                                                                                                    & Faster RCNN-Res50            & 38.4                                        & 21.5                                  & 42.1                                  & 50.3                                  & 52.0                                        & 32.6                                  & 55.8                                  & \multicolumn{1}{l}{66.1}              \\
                                                                                                    & FKD \cite{zhang2020improve}                          & 41.5 (+3.1)                                  & 23.5                                  & 45.0                                  & 55.3                                  & 54.4 (+2.4)                                  & 34.0                                  & 58.2                                  & \multicolumn{1}{l}{69.9}              \\
                                                                                                    & FGD \cite{yang2022focal}                          & 42.0 (+3.6)                                  & 23.8                                  & 46.4                                  & 55.5                                  & 55.4 (+3.4)                                  & \textbf{35.5}                                     & 60.0                        & 70.0                                  \\
                                                                                                    & MGD \cite{yang2022masked}                          & 42.1 (+3.7)                                  & 23.7                                  & 46.4                                  & 56.1                                  & 55.5 (+3.5)                                  & 35.4                         & 
                                                                             60.0                        & 70.5                                  \\
\multirow{-5}{*}{\begin{tabular}[c]{@{}c@{}}Cascade\\ Mask RCNN\\ ResNeXt101\\ (47.3)\end{tabular}} & \cellcolor[HTML]{EFEFEF}AMD (ours) & \cellcolor[HTML]{EFEFEF}\textbf{42.4 (+4.0)} & \cellcolor[HTML]{EFEFEF}\textbf{24.1} & \cellcolor[HTML]{EFEFEF}\textbf{46.5} & \cellcolor[HTML]{EFEFEF}\textbf{56.2} & \cellcolor[HTML]{EFEFEF}\textbf{55.8 (+3.8)} & \cellcolor[HTML]{EFEFEF}35.3          & \cellcolor[HTML]{EFEFEF}\textbf{60.0} & \cellcolor[HTML]{EFEFEF}\textbf{70.8} \\ \hline

\end{tabular}
\label{results}
}
\end{table*}

We also conduct a series of ablation studies to explore the effects of individual components on the performance of our AMD framework. In implementation, all the experiments are conducted on a server with one RTX3090 GPU using MMdetection toolbox \cite{chen2019mmdetection} and Pytorch framework \cite{paszke2019pytorch}. Besides, the hyper-parameters are empirically set to $\left\{\alpha=2.5 \times 10^{-7}, \lambda=1, \mathcal{T}=0.5\right\}$ and $\left\{\alpha=4 \times 10^{-6}, \lambda=1.2, \mathcal{T}=0.5\right\}$ for the one-stage models and the two-stage models respectively. During the training process, SGD optimizer is used for training all the detectors within 24 epochs. Meanwhile, momentum is set as 0.9 whilst weight decay is set to 0.0001. Moreover, single-scale
training strategy is utilized in our experiments.

\subsection{Results}
In our comparative studies, we carry out three groups of experiments to evaluate different distillation methods with the three popular detectors involved. The corresponding experimental results are shown in Table \ref{results}. 
\begin{table*}[htb]
\centering
\caption{Ablation studies using RetinaNet \cite{lin2017focal} framework for both the teacher and the student. The backbone of the teacher network is ResNeXt-101 whilst its student counterpart is ResNet-50. Ada-Mask and Ada-channel respectively denote spatially adaptive masking and channel adaptive clue generation module. They constitute two main components in our proposed AMD model.}
\resizebox{0.6\textwidth}{!}{%
\begin{tabular}{cc|ccccccc}
\toprule
\multicolumn{1}{c}{\multirow{2}{*}{Ada-Mask}} &
  \multicolumn{1}{c|}{\multirow{2}{*}{Ada-Channel}} &
  \multicolumn{6}{c}{Student: RetinaNet + Res50} \\ \cmidrule(l){3-8} 
\multicolumn{1}{c}{} &
  \multicolumn{1}{c|}{} &
  \multicolumn{1}{c|}{$AP^{b}$} &
  \multicolumn{1}{c|}{$AP^{b}_{50}$} &
  \multicolumn{1}{c|}{$AP^{b}_{75}$} &
  \multicolumn{1}{c|}{$AP_{S}$} &
  \multicolumn{1}{c|}{$AP_{M}$} &
  \multicolumn{1}{c}{$AP_{L}$} \\ \midrule \rowcolor[gray]{0.85}
\ding{52} & \ding{52} & \multicolumn{1}{c|}{\textbf{41.3}} & \multicolumn{1}{c|}{\textbf{61.0}} & \multicolumn{1}{c|}{\textbf{44.1}} & \multicolumn{1}{c|}{\textbf{23.9}} & \multicolumn{1}{c|}{\textbf{45.4}} & \multicolumn{1}{c}{\textbf{55.7}} \\
 & \ding{52} & \multicolumn{1}{c|}{41.0} & \multicolumn{1}{c|}{61.0} & \multicolumn{1}{c|}{43.8} & \multicolumn{1}{c|}{23.7} & \multicolumn{1}{c|}{45.3} & 55.6                     \\
 \ding{52} &  &  \multicolumn{1}{c|}{41.2} & \multicolumn{1}{c|}{60.8} & \multicolumn{1}{c|}{44.0} & \multicolumn{1}{c|}{23.4} & \multicolumn{1}{c|}{45.2} & 55.6                     \\ \bottomrule
\end{tabular}%
}
\label{ab_retina}
\end{table*}

\begin{table*}[htb]
\centering
\caption{Ablation studies using RepPoint \cite{yang2019reppoints} framework for both the teacher and the student. ResNeXt-101 and ResNet-50 are respective backbones.}
\resizebox{0.6\textwidth}{!}{%
\begin{tabular}{cc|ccccccc}
\toprule
\multicolumn{1}{c}{\multirow{2}{*}{Ada-Mask}} &
  \multicolumn{1}{c|}{\multirow{2}{*}{Ada-Channel}} &
  \multicolumn{6}{c}{Student: RepPoint + Res50} \\ \cmidrule(l){3-8} 
\multicolumn{1}{c}{} &
  \multicolumn{1}{c|}{} &
  \multicolumn{1}{c|}{$AP^{b}$} &
  \multicolumn{1}{c|}{$AP^{b}_{50}$} &
  \multicolumn{1}{c|}{$AP^{b}_{75}$} &
  \multicolumn{1}{c|}{$AP_{S}$} &
  \multicolumn{1}{c|}{$AP_{M}$} &
  \multicolumn{1}{c}{$AP_{L}$} \\ \midrule \rowcolor[gray]{0.85}
\ding{52} & \ding{52} &  \multicolumn{1}{c|}{\textbf{42.7}} & \multicolumn{1}{c|}{\textbf{63.5}} & \multicolumn{1}{c|}{\textbf{46.5}} & \multicolumn{1}{c|}{\textbf{24.8}} & \multicolumn{1}{c|}{\textbf{46.5}} & \multicolumn{1}{c}{\textbf{56.3}} \\
 & \ding{52} & \multicolumn{1}{c|}{42.4} & \multicolumn{1}{c|}{63.2} & \multicolumn{1}{c|}{46.4} & \multicolumn{1}{c|}{24.6} & \multicolumn{1}{c|}{46.5} & 56.1                     \\
\ding{52} &  &  \multicolumn{1}{c|}{42.4} & \multicolumn{1}{c|}{63.3} & \multicolumn{1}{c|}{46.2} & \multicolumn{1}{c|}{24.4} & \multicolumn{1}{c|}{46.3} & 56.0                     \\ \bottomrule
\end{tabular}%
}
\label{ab_reppoint}
\end{table*}

\begin{table*}[htb]
\centering
\caption{Ablation studies in a cross-framework scenario. The Cascade Mask-RCNN \cite{cai2019cascade} is employed for the teacher framework, while the Faster R-CNN is for the student counterpart.}
\resizebox{0.6\textwidth}{!}{%
\begin{tabular}{cc|ccccccc}
\toprule
\multicolumn{1}{c}{\multirow{2}{*}{Ada-Mask}} &
  \multicolumn{1}{c|}{\multirow{2}{*}{Ada-Channel}} &
  \multicolumn{6}{c}{Student: Faster-RCNN + Res50} \\ \cmidrule(l){3-8} 
\multicolumn{1}{c}{} &
  \multicolumn{1}{c|}{} &
  \multicolumn{1}{c|}{$AP^{b}$} &
  \multicolumn{1}{c|}{$AP^{b}_{50}$} &
  \multicolumn{1}{c|}{$AP^{b}_{75}$} &
  \multicolumn{1}{c|}{$AP_{S}$} &
  \multicolumn{1}{c|}{$AP_{M}$} &
  \multicolumn{1}{c}{$AP_{L}$} \\ \midrule \rowcolor[gray]{0.85}
\ding{52} & \ding{52} &  \multicolumn{1}{c|}{\textbf{42.4}} & \multicolumn{1}{c|}{\textbf{63.1}} & \multicolumn{1}{c|}{\textbf{46.2}} & \multicolumn{1}{c|}{\textbf{24.1}} & \multicolumn{1}{c|}{46.5} & \multicolumn{1}{c}{56.2} \\
 & \ding{52} &  \multicolumn{1}{c|}{42.1} & \multicolumn{1}{c|}{62.8} & \multicolumn{1}{c|}{46.0} & \multicolumn{1}{c|}{23.8} & \multicolumn{1}{c|}{46.4} &  \textbf{56.3}                    \\
\ding{52} &  &  \multicolumn{1}{c|}{42.3} & \multicolumn{1}{c|}{63.0} & \multicolumn{1}{c|}{46.2} & \multicolumn{1}{c|}{23.6} & \multicolumn{1}{c|}{\textbf{46.6}} & 56.3                     \\ \bottomrule
\end{tabular}%
}
\label{ab_maskrcnn}
\end{table*}

In the first group of experiments, RetinaNet is used as the detection framework for both the teacher and the student. The corresponding experimental results demonstrate that our distillation method provides significant performance boosts of 3.9$\%$ in mAP over the baseline student network by reporting the highest accuracy at 41.3$\%$. This result consistently outperforms the state-of-the-art methods FGD and MGD by 0.6$\%$ and 0.3$\%$, while it even surpasses the teacher model achieving 41.0$\%$ mAP. Similar performance improvement can also be observed with respect to mAR metric. The experimental setting in the second group is analogous to the first one except that the RetinaNet framework is replaced with RepPoints. Consistent with the results in the first group, dramatic performance gains of 4.1$\%$ in mAP and 3.7$\%$ in mAR are reported, and similar performance superiority to the competing distillation methods is also demonstrated. The results reveal that our method can adaptively learn more important information from the teacher and significantly contribute to the improvement of the student model.

To further assess the generalization capability of our proposed method, we make use of different detection frameworks for the teacher and student models. To be specific, the more powerful detector Cascade Mask-RCNN is used as the teacher network while the Faster-RCNN for the student model. As shown in Table \ref{results}, our method boosts the baseline student model from 38.4$\%$ to 42.4$\%$ in mAP and from 52.0$\%$ to 55.8$\%$ in mAR, outperforming MGD 0.3$\%$ both in mAP and mAR. It sufficiently suggests our method is independent of the specific detector and shows consistent advantages in cross-framework scenarios.  

\subsection{Ablation Study}
In this section, we conduct extensive ablation experiments to explore the effect of different configurations on the proposed AMD. Consistent with the above setting, the ablation experiments with different configurations are conducted based on the three popular detectors, \textit{i.e.,} RetinaNet, Faster-RCNN, and RepPoint. 

As shown in Table \ref{ab_retina}, when RetinaNet is used for the detection framework for both the teacher and the student, we explore two primary modules in our AMD model, namely the spatially adaptive masking (Ada-Mask) and the channel adaptive clues generation (Ada-Channel). It is observed that the complete AMD model including both the Ada-Mask and Ada-Channel components achieves the best results. Furthermore, when we remove either component, there is a clear performance drop in particular in the small-object detection scenario (0.3\%$\downarrow$ w/o Ada-Mask and 0.5\%$\downarrow$ w/o Ada-Channel). This implies that our AMD method can improve object-awareness capability which is crucial for dense prediction tasks.

When the RetinaNet is replaced with the RepPoint, similar results can be obtained. As displayed in Table \ref{ab_reppoint}, both the Ada-Mask and Ada-Channel components play critical roles in our AMD model. Specifically, single Ada-Mask module reports 24.4\%, 46.3\% and 56.0\% in $AP_{S}$, $AP_{M}$ and $AP_{L}$ scores. With the help of additional channel adaptive clues, further performance gains of 0.4\%, 0.2\% and 0.3\% are reported for the respective metrics.

Furthermore, we also perform ablation studies in cross-framework scenarios. Specifically, the Cascade Mask-RCNN is used as the teacher network, while the Faster-RCNN as the student counterpart. As shown in Table \ref{ab_maskrcnn}, the complete AMD model achieves the highest accuracy. In particular, the highest $AP_{S}$ score 24.1\% is reported, outperforming the other settings w/o either Ada-Mask or Ada-Channel. This indicates that our AMD model benefits small-object detection with improved object-awareness capability.

\begin{table}[htb]
\caption{Comparison of different generation blocks. For MBConv \cite{tan2019efficientnet}, we use $5\times 5$ depthwise convolution.}
\resizebox{\columnwidth}{!}{%
\begin{tabular}{@{}clll@{}}
\toprule
\multicolumn{4}{c}{Student: RetinaNet-Res50}                                                                                       \\ \midrule
\multicolumn{1}{c|}{Generation Block} & \multicolumn{1}{l|}{MBConv} & \multicolumn{1}{l|}{$3\times 3$ Dense Conv * 1} & $3\times 3$ Dense Conv * 2 \\ \midrule
\multicolumn{1}{c|}{mAP}              & \multicolumn{1}{c|}{41.0}       & \multicolumn{1}{c|}{41.2}                   &    \multicolumn{1}{c}{\textbf{41.3}}               \\ \bottomrule
\end{tabular}%
}
\label{ab_generation}
\end{table}

In addition to the above ablation studies, we also discuss the effect of different generation blocks on the performance of our method. As illustrated in Table \ref{ab_generation}, three different generation blocks are compared within the RetinaNet framework. The results reveal that a slightly inferior performance is reported by the advanced MBConv \cite{tan2019efficientnet}. In contrast, a better result is achieved by simply stacking two vanilla convolutional layers. We assume that the channel adaptive clues learned from the teacher network is not compatible with MBConv block, because MBConv somewhat encodes the channel clues from the student model. This incompatibility results from the difference of the channel clues between the teacher and the student network.

\begin{table}[htb]
\centering
\caption{Comparison of different locations of Ada-channel. After and Within denote that we apply the channel adaptive clues after the generation block and between the two convolution layers, respectively.}
\resizebox{0.25\textwidth}{!}{%
\begin{tabular}{@{}ccc@{}}
\toprule
\multicolumn{3}{c}{Student: Faster-RCNN + Res50}                    \\ \midrule
\multicolumn{1}{c|}{Location} & \multicolumn{1}{c|}{After} & Within \\
\multicolumn{1}{c|}{mAP}      & \multicolumn{1}{c|}{\textbf{42.4}}  & 42.2   \\ \bottomrule
\end{tabular}%
}
\label{ab_location}
\end{table}
To gain a deeper insight into the effect of the Ada-Channel module on feature generation, we explore the following two cases with Cascade Mask-RCNN and Faster-RCNN respectively used as the teacher and the student. In the first case, Ada-Channel follows the generation block, and the two components function separately. In the other case, Ada-Channel is embedded within two consecutive convolution layers of the generation block, which implies that two modules are coupled. As shown in Table \ref{ab_location}, decoupling the two components brings an improvement of 0.2\% in mAP, suggesting that the generation process working on the masked feature of the student is repulsive with other exotic clues, even the informative ones.

\subsection{Parameter Analysis}

\begin{figure}[htb]
\centerline{\includegraphics[width=0.4\textwidth]{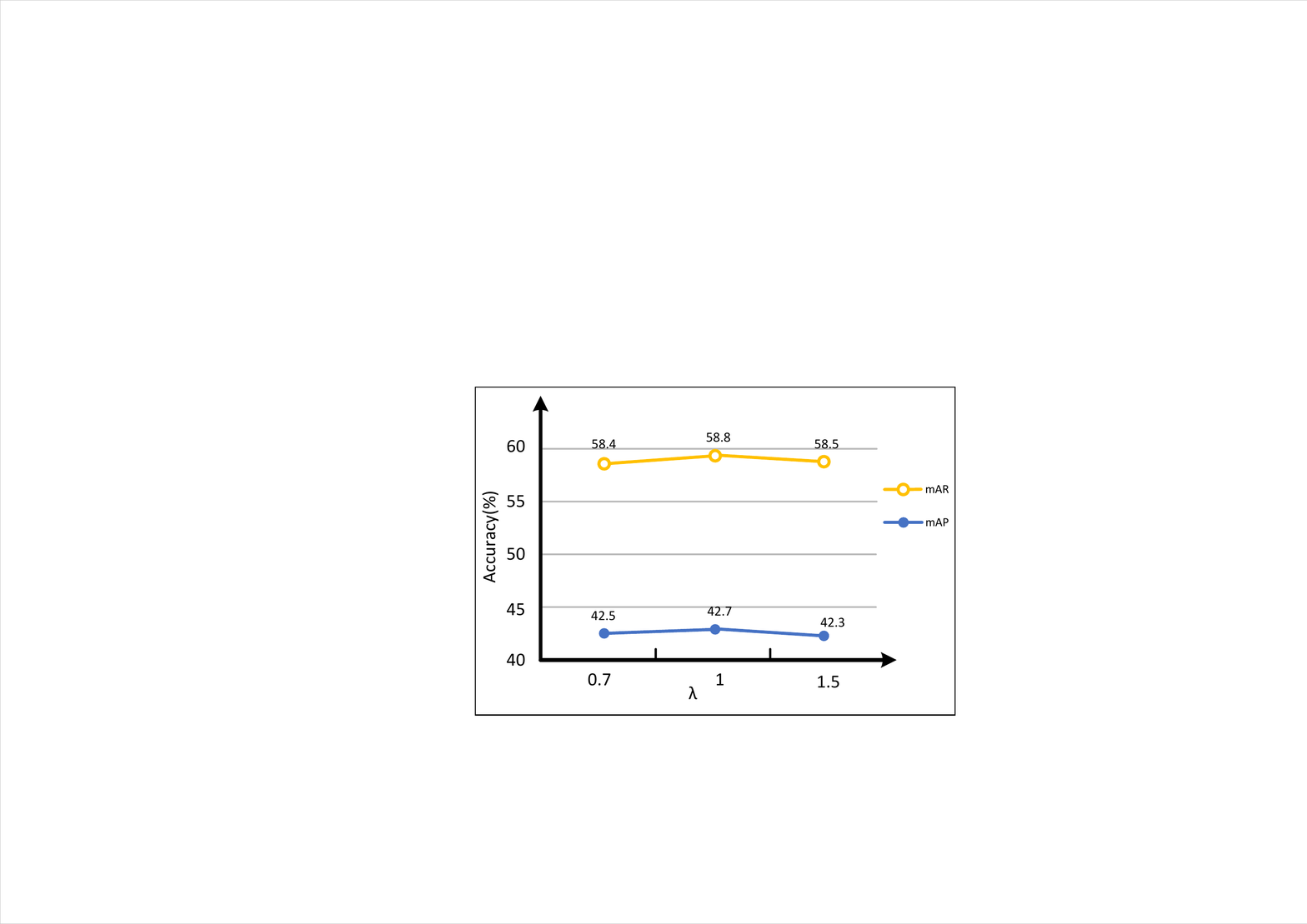}}
\caption{Parameter $\lambda$ analysis on one-stage RepPoints framework.}
\label{zhexian}
\end{figure}

In our AMD method, the hyper-parameter $\lambda$ in Eq. \ref{M} controls the coverage of feature mask. A larger $\lambda$ value indicates that only the points with higher attention scores of the teacher model are masked, and most of the pixel points are in the object-specific ground-truth region. In contrast, it is likely that masked points appear in the background region when decreasing $\lambda$. In our experiments, we discuss the effect of $\lambda$ using RepPoints as the detection framework. It is observed from Fig. \ref{zhexian} that the highest mAP 42.7\% and mAR 58.8\% are reported when $\lambda=1.0$, suggesting it helps the model to better compromise between encoding low-score and high-score regions.

%% file: Conclu.tex
\section{Conclusion}
In this paper, we focus on the topic of feature-based masked distillation and propose spatial-channel adaptive masked distillation termed AMD for object detection. On the one hand, we perform spatially adaptive feature masking to encode the region-specific importance, such that more important and expressive features can be learned from the teacher network. On the other hand, to improve the object-awareness capability, we utilize the simple and efficient SE block to generate informative channel-adaptive clues for the student model. Extensive experiments demonstrate the superiority and effectiveness of our method, showing that the proposed AMD model not only significantly boosts the performance of the baseline student model but also outperforms the other state-of-the-art distillation approaches.

In our proposed AMD, the spatial attention map generated from the feature of the teacher model lacks information interaction. Our future work will focus on exploring alternative strategies to enhance the interaction among different locations on the attention map.